\documentclass[11pt,a4paper]{article}
\usepackage[margin=1in]{geometry}
\usepackage{fontspec}
\usepackage{amsmath,graphicx,booktabs,tabularx,array,float,placeins,setspace}
\usepackage[authoryear,round]{natbib}
\usepackage[hidelinks,breaklinks]{hyperref}
\usepackage{url}
\usepackage[font=small,labelfont=bf]{caption}
\date{}
\title{Social Influence and the Allocation of Scientific Attention in AI Populations}
\author{Maxim Chupilkin\\[3pt]{\small Department of Politics and International Relations}\\{\small University of Oxford}\\{\small \texttt{maxim.chupilkin@politics.ox.ac.uk}}}
\begin{document}
\maketitle

\begin{abstract}AI systems are becoming participants in the evaluation and use of scientific research. They encounter citation counts, download statistics and lists of popular articles developed around human readers, but the collective consequences of these signals for artificial readers remain uncertain. This paper adapts the Music Lab design to a market for academic attention. In the first experiment, 1,000 AI agents choose papers from the titles and abstracts of all 114 regular research articles published in the American Economic Review in 2025. The experiment has five independent-choice communities and five social-influence communities, each with 100 sequential agents. Only agents in the social-influence condition observe earlier selections within their community. Agents may select any number of papers. Social-information communities select 17.2 percent fewer papers per agent, concentrate their choices more heavily, and collectively cover 73 papers, compared with 90 independently. Between-community variation is greater under social information. In a second experiment with 200 agents across twenty social communities, randomly assigning papers five initial selections raises their subsequent selection rate by 45.55 percentage points (95\% CI: 41.20 to 49.90). Choices have modest correspondence with external citations and little correspondence with download counts. The results show how a simple information rule shapes the volume, breadth and distribution of scientific attention in an artificial population.\end{abstract}

\FloatBarrier\section*{1. Introduction}

How does information about others' choices shape which research receives attention from AI systems? Scientific papers are encountered alongside citation counts, download statistics and lists of popular articles. These signals help human readers navigate a growing literature; AI systems now enter the same information environment. Recent studies introduce The AI Scientist, which generates ideas, runs experiments and writes manuscripts, and Co-Scientist and Robin, which generate and refine scientific hypotheses (\citealp{lu2026}; \citealp{gottweis2026}; \citealp{ghareeb2026}). Paper2Agent turns papers into interactive research agents (\citealp{miao2026}), while the American Economic Association's partnership with Refine introduces AI-assisted technical verification into journal workflows (\citealp{aea2026}). These developments make it increasingly important to understand how AI systems select research and how social information shapes their collective attention.

This paper studies these choices by adapting Salganik, Dodds and Watts's Music Lab experiment to a market for academic attention (\citealp{salganik2006}). The experiment comprises five independent-choice communities and five social-influence communities, each with 100 sequential agents using GPT-5.6 Sol. Every agent receives the titles and abstracts of all 114 regular research articles published in the American Economic Review in 2025, presented in a random order, and chooses which papers to read in full. Agents may select any number, including none. Independent agents see no information about others' choices; social agents see the cumulative number of earlier selections for each paper within their community. Holding the corpus constant across separate communities makes it possible to study both concentration within communities and differences between them. A second experiment randomly assigns early popularity to papers in twenty additional social communities of ten agents each, testing whether an arbitrary head start changes subsequent attention.

Visible popularity can shape collective attention in three ways. It can reinforce preferences that agents already share, concentrating choices around common favourites. It can also amplify arbitrary early differences, directing otherwise identical communities towards different papers. Models of herding and informational cascades explain how earlier choices can influence later decisions (\citealp{banerjee1992}; \citealp{bikhchandani1992}). Finally, popularity may change how selective agents are: evidence of others' interest could become a condition for choosing a paper, or draw attention to additional works. These possibilities motivate examining selection volume, concentration and differences across communities together. The randomised experiment tests whether an initial popularity advantage changes subsequent choices of the same papers.

Social information is associated with fewer selections, narrower collective coverage and greater concentration. Independent agents select an average of 17.29 papers, compared with 14.32 in social communities, a reduction of 17.2 percent. Across the five communities in each condition, 90 papers receive at least one independent selection, compared with 73 under social information. Mean within-community Gini rises from 0.753 to 0.844, and all five social communities have lower selection volume and greater concentration than all five independent communities. Selection shares also differ more across social communities. In the second experiment, a randomly assigned head start raises subsequent selection rates from 14.45 to 60.00 percent, a 45.55-percentage-point difference (95\% CI: 41.20 to 49.90). Choices have modest correspondence with external citations and little correspondence with download counts.

The paper speaks to research on attention, cumulative advantage and artificial societies. The economics of attention examines how consideration is allocated across competing objects (\citealp{loewenstein2025}); research on citation indexing, scientific recognition and author status shows how information and reputation shape the visibility of ideas (\citealp{garfield1955}; \citealp{merton1968}; \citealp{simcoe2011}). Experiments in cultural markets and online platforms demonstrate that manipulated popularity and early advantages can redirect subsequent choices (\citealp{salganik2008}; \citealp{muchnik2013}; \citealp{vanderijt2014}). In artificial populations, Ashery, Aiello and Baronchelli study emergent conventions and collective bias (\citealp{ashery2025}), Wu et al. examine popularity and choice (\citealp{wu2025}), and CiteAgent studies citation-network formation (\citealp{ji2026}). Recent work also links AI use to a narrowing of scientific topics (\citealp{hao2026}), while a preprint studies concentration in ideas generated by research agents (\citealp{tang2026}). This paper examines the institutional allocation of attention to existing work: how visible choices organise reading selections across AI communities.

The main contribution is to measure how an information rule shapes the amount, breadth and distribution of attention in an artificial society. Unrestricted choice allows total demand to respond to social information alongside concentration and variation across communities. The resulting evidence shows that narrower collective attention can emerge even when every agent receives every abstract and can choose freely. Randomised initial popularity further establishes that arbitrary early advantages can redirect this attention across the same papers. AI agents are the population of interest: the experiment uses a social-science design to study their collective behaviour and the institutions governing it. Academic papers provide a setting in which these institutional questions matter for the growing use of AI in research.

Section 2 describes the corpus, experimental design and outcome measures. Section 3 presents the first experiment's results. Section 4 tests randomly assigned early popularity, Section 5 compares selections with external measures of scholarly attention, and Section 6 discusses the implications for artificial research communities.

\FloatBarrier\section*{2. Experimental design and measurement}

\subsection*{2.1. Corpus and population}

AER is especially useful for studying the allocation of scientific attention. It is the American Economic Association's flagship journal and a \href{https://www.aeaweb.org/journals/aer}{general-interest outlet} covering a broad range of economics topics. AER also belongs to the discipline's "top five" journals, whose publications strongly influence tenure decisions in leading US economics departments (\citealp{heckman2020}). A single annual cohort combines variation in subjects and methods with a common venue and editorial selection process. Agents choose among papers that have already passed the same journal's publication threshold, while venue and year are held constant. The corpus is also small enough to present every abstract to every agent.

The setting connects the experiment to observable scholarly activity. All 114 papers can be matched to OpenAlex citation counts and RePEc/LogEc download-click statistics, permitting comparison with external attention. Economics also provides concrete applications of AI to research: Korinek documents uses in literature research, analysis and writing (\citealp{korinek2023}), and the AEA's Refine partnership brings AI-assisted technical verification into journal workflows (\citealp{aea2026}). Together, the breadth of the corpus, common publication setting and available attention measures make AER a useful setting for this line of work.

The corpus comprises all 114 regular research articles in the twelve issues of AER volume 115, published in 2025. Article titles, abstracts, issue metadata and DOIs were collected from the journal's official website. Front matter, the auditor report, Nobel lectures, presidential addresses and comments were excluded. The corpus is consequently a complete cohort under an explicit document-type restriction, not a random sample of all economics research. Source HTML, cleaned records and content fingerprints are retained. No abstract was rewritten to standardise its appeal.

Every request contains a paper ID, title and complete abstract for each item. Author names, publication dates, the journal label and external attention measures are omitted as separate fields. Abstracts are presented without modification.

The experiment uses GPT-5.6 Sol through OpenAI's direct Responses API. The choice of the GPT family is motivated by the widespread use of OpenAI's systems: a six-country survey found ChatGPT to be the most widely used generative AI system, with 22 percent reporting weekly use (\citealp{simon2025}). Each decision is a new context containing only the task and the current paper menu. Agents do not have tools, persistent personal histories or access to full texts. Five independent communities and five social-information communities each contain 100 sequential decisions. The ten communities can run concurrently, but each social history is updated only after the preceding decision in that community. There is no exchange of counts across communities.

Each agent is a fresh model invocation with one choice opportunity. The population shares a model and task instruction; no demographic or occupational personas are assigned. This design studies the collective behaviour produced by a common decision system under different information rules.

\subsection*{2.2. Task and social information}

The main instruction is: "Based on their titles and abstracts, decide which papers you would choose to read in full. You may select any number, including none. Select a paper only if you would actually choose to read it, rather than merely regard it as potentially interesting." The response is an unordered list of distinct paper IDs. Omitted papers are coded as unselected. There is no requirement to justify decisions, rate interest or select a minimum number. Appendix A reproduces the complete system instruction and condition-specific text.

Social agents receive a previous-selections field beside every paper. It reports the number of earlier agents within their own community that selected the paper to read in full. All counts start at zero. The accompanying text explains that earlier agents could choose any number, including none, and that the counts belong only to the current community. Independent agents receive the statement that no information about other agents' selections is provided. The treatment therefore consists of both the count display and its explanation.

Paper order is shuffled separately for each community and visit using a recorded deterministic schedule. Corresponding community and visit indices use the same order across conditions. Calls remain separate model draws. Condition schedules are constructed explicitly, with presentation order matched across them.

The first social agent in each community sees only zeros and supplies the first positive counts available to the second agent. Comparing initial choices therefore describes behaviour before positive social history has accumulated.

\begin{figure}[H]\centering\includegraphics[width=\linewidth]{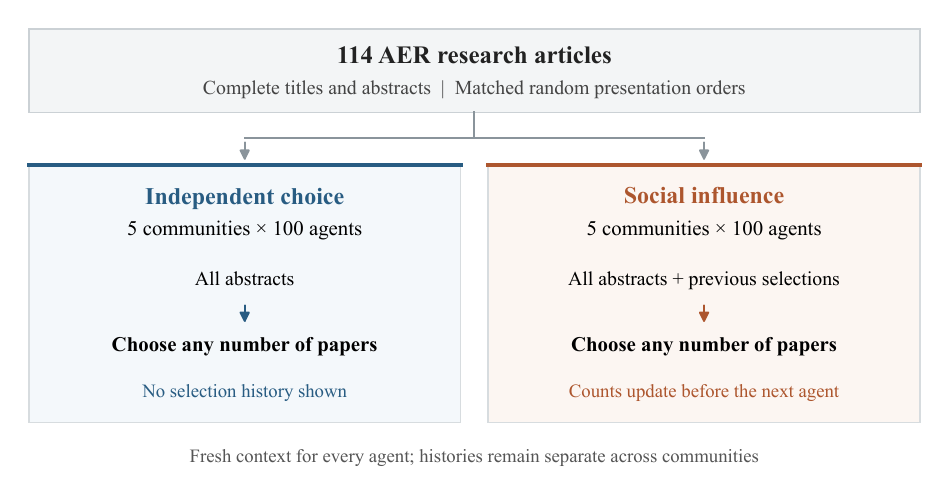}\caption*{Figure 1. Experimental design.}

{\footnotesize\setstretch{1.1}\raggedright \textit{Note: Both conditions use the same corpus and matched presentation orders. Only the social condition displays cumulative selections from earlier agents in the same community; its counts begin at zero.}\par}

\end{figure}

\subsection*{2.3. Outcomes and denominators}

Let y(i,w,t) equal one if agent t in world w selects paper i, and zero otherwise. Individual selection volume is the sum of these indicators across the corpus. Final paper counts add the indicators over 100 visits. Within-world coverage counts papers with at least one selection. Arm-wide coverage takes the union over the five communities in the same condition. This distinction matters: an arm can collectively cover a paper even if most of its individual worlds never select it.

Concentration is measured by the Gini coefficient of the final count vector, including all zero-count papers. For P papers, counts c(i,w) and total selections C(w), the coefficient is the sum of all pairwise absolute count differences divided by 2P times C(w). The reported arm statistic is the mean of the five world-level coefficients, not the Gini of pooled counts. Empty worlds would have an undefined coefficient under this convention; none occurs in the completed experiment.

Between-world variation is based on shares s(i,w) = c(i,w)/C(w). For each pair of worlds, total variation is half the sum across papers of the absolute difference in their shares. The arm-level statistic averages this distance over the ten distinct pairs. The Music Lab-style mean absolute share difference is also retained in the analytical data; for a fixed corpus it equals twice total variation divided by the number of papers. The share denominator uses each world's actual selections, which is necessary when demand is unrestricted.

The three outcomes distinguish the breadth, concentration and variability of attention. Total variation uses normalised shares, allowing comparisons when selection volume differs. The ten pairwise distances within a condition share the same five communities. Uncertainty is therefore estimated at the community level.

\subsection*{2.4. Uncertainty}

Figures for the first experiment report approximate 95\% confidence intervals across the five communities in each condition. Mean outcomes use the community standard error and a Student t critical value with four degrees of freedom. For mean pairwise total variation, a delete-one-community jackknife accounts for dependence among the ten distances. Trajectory bands are pointwise. These intervals describe repeated-community uncertainty conditional on the corpus, model and protocol; they are not randomisation-based intervals. Appendix A gives the calculations.

\subsection*{2.5. Collection and validation}

The experiment yielded 1,000 valid decisions. The saved responses returned the model identifier gpt-5.6-sol; no immutable dated model version was supplied. Reasoning effort was set to low, the output limit to 2,048 tokens, and temperature was omitted so the provider default applied. Every response was checked for completion, valid paper IDs and absence of duplicate IDs. A separate verifier reconstructed every prompt and the before-and-after community counts. The recorded input includes the full abstract set at every visit.

Saved requests and responses support reconstruction of every choice history. Appendix A gives the exact instructions and statistical methods; full API settings are retained in the replication records. Figures and tables are generated from frozen records without additional model calls.

\FloatBarrier\section*{3. Results}

\subsection*{3.1. Unrestricted choice remains selective}

Unrestricted choice did not produce selection of the whole corpus. Independent agents selected between 10 and 29 papers, while social-information agents selected between 8 and 22. No agent selected none. The main task therefore produces variation in expressed demand while avoiding both degenerate extremes in the observed calls. The object being allocated is a planned reading list: the experiment does not require agents to bear the time cost of reading the selected works.

Table 1 reports the main outcomes. Mean selection volume is 17.29 in the independent condition and 14.32 under social information, a descriptive reduction of 17.2 percent. Across all visits this corresponds to 8646 and 7161 selections, respectively. The difference is visible at the community level rather than depending on one particularly short list. Figure 2 shows every community's mean choice count and Gini, retaining the small number of replications in the display.

\begin{table}[H]\centering\caption*{\textbf{Table 1. Unrestricted choices among 114 papers}}\begingroup\small\setstretch{1.1}\setlength{\tabcolsep}{4pt}\begin{tabularx}{\linewidth}{Xrr}\toprule

\textbf{Outcome} & \textbf{Independent} & \textbf{Social} \\\midrule

Total selections & 8646 & 7161 \\

Mean selections per agent & 17.29 & 14.32 \\

Median selections per agent & 17.0 & 14.0 \\

Distinct papers across 5 worlds & 90 & 73 \\

Mean within-world Gini & 0.753 & 0.844 \\

Mean pairwise total variation & 0.073 & 0.106 \\

\bottomrule\end{tabularx}\endgroup

{\footnotesize\setstretch{1.1}\raggedright \textit{Note: 5 communities and 500 agents per condition. Coverage is the union across communities; Gini and total variation use world-level distributions.}\par}

\end{table}

\begin{figure}[H]\centering\includegraphics[width=\linewidth]{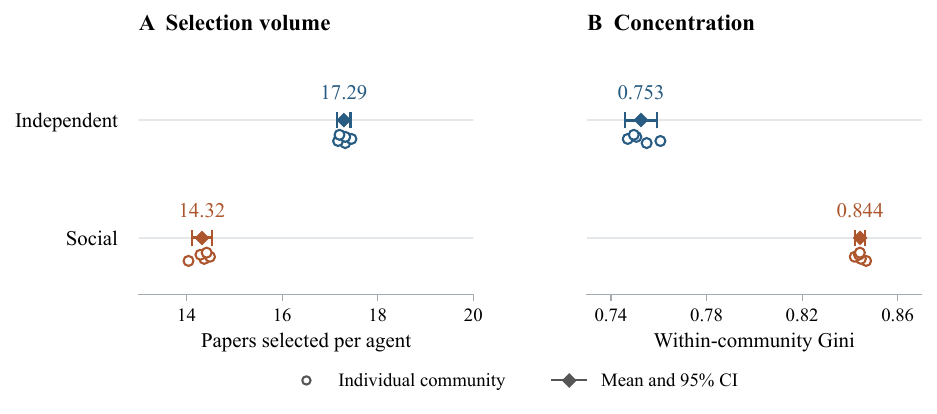}\caption*{Figure 2. Social communities select fewer papers and concentrate attention.}

{\footnotesize\setstretch{1.1}\raggedright \textit{Note: Open circles show the five communities of 100 agents per condition, offset vertically for readability. Diamonds and labels show condition means; whiskers are approximate 95\% community-level t intervals (4 degrees of freedom). Gini includes all 114 papers. Both horizontal axes are truncated.}\par}

\end{figure}

The social communities choose shorter lists despite receiving the same abstracts and freedom to select papers. One interpretation is that a popularity display induces a more selective decision threshold. Another is that the model infers an implicit norm from the community context. The instruction asks agents to select papers they would actually read, so the lower volume reflects a difference in expressed selectivity. At the first visit, when all social counts are zero, the mean difference is not statistically significant (social minus independent: -2.00 papers; 95\% CI: -6.97 to 2.97; paired t test, p = 0.326).

\subsection*{3.2. Collective coverage narrows and concentration rises}

Across the five independent communities, 90 of the 114 papers receive at least one selection, compared with 73 across social communities. Within a single independent world, coverage ranges from 68 to 79 papers; the corresponding social range is 52 to 55. Social communities thus share a narrower collective menu without approaching complete agreement on an identical list. Both within-world and arm-wide definitions point in the same direction.

The mean Gini is 0.753 for independent worlds and 0.844 for social worlds. Every social-world coefficient exceeds every independent-world coefficient. Mean within-world coverage is 74.8 papers independently and 53.4 socially. Visible social information is thus associated with both a smaller number of selections and a more concentrated distribution of those selections.

Figure 3 traces cumulative coverage over visits and over realised selections. Panel A shows condition means with 95\% confidence bands and individual community trajectories. Panel B plots coverage against the number of selections actually made in each community, showing how the breadth of attention develops alongside its volume.

\begin{figure}[H]\centering\includegraphics[width=\linewidth]{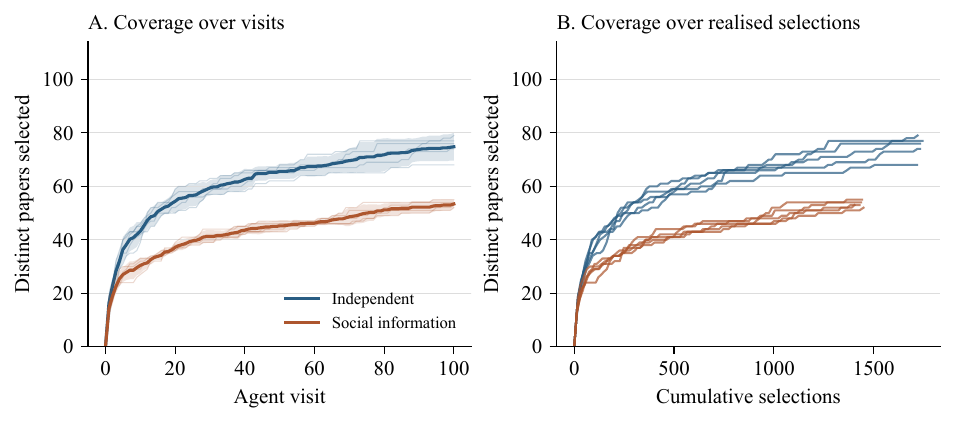}\caption*{Figure 3. Cumulative coverage expands more slowly in social worlds.}

{\footnotesize\setstretch{1.1}\raggedright \textit{Note: Panel A shows condition means and pointwise 95\% community-level t intervals; faint lines show individual worlds. Panel B shows individual worlds against realised cumulative selections. Selection volume is endogenous.}\par}

\end{figure}

Shared preferences remain prominent. Some papers attract almost universal selections in both conditions. One paper is selected by every agent in both conditions. The heatmap in Figure 4 displays all papers receiving any selection in the unrestricted experiment, ordered by pooled selection frequency with paper ID breaking ties. The same row order applies to every world. Broadly shared favourites coexist with a tail of papers selected only occasionally or only in independent communities. This structure makes it plausible that part of social concentration reflects reinforcement of common model preferences rather than the creation of entirely different leading works.

\begin{figure}[H]\centering\includegraphics[width=\linewidth]{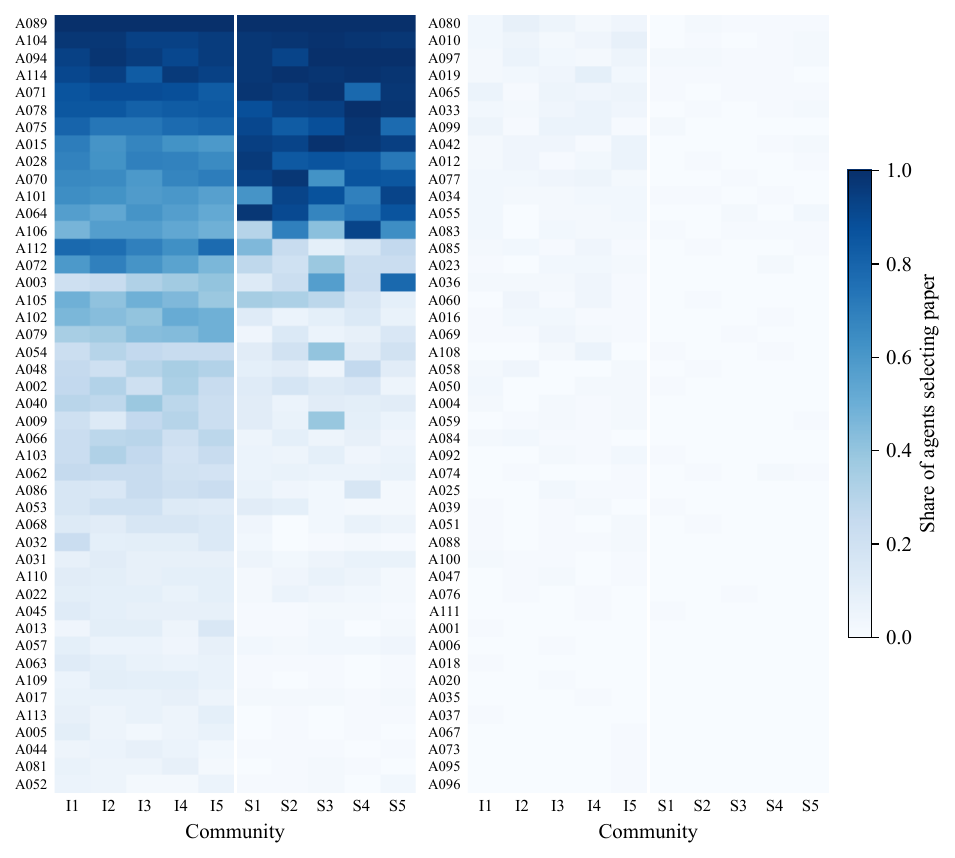}\caption*{Figure 4. Common favourites coexist with differences across histories.}

{\footnotesize\setstretch{1.1}\raggedright \textit{Note: All papers selected in either condition, ordered by combined frequency, continuing from the left panel to the right. I denotes independent and S social worlds. Colour shows selections divided by 100 agents. Corpus IDs map to titles and DOIs in the exhibit data; all-zero papers are omitted here but included in all statistics.}\par}

\end{figure}

\subsection*{3.3. Shared favourites, persistent early advantages}

The mechanism question is whether social information narrows attention around papers that agents already favour, or allows different early histories to shape later choices. The data contain evidence of both. Aggregate paper rankings are highly similar across conditions: their Spearman correlation is 0.953 (Figure 5A). Yet 80 percent of selections fall on an average of only 13.2 papers in a social community, compared with 22.6 in an independent community. Social information concentrates attention around a smaller core while largely preserving the underlying ranking.

To examine the role of history, we compare each paper's selection rate in the first ten visits with its rate in visits 11-100. The analysis removes differences in overall paper popularity and community selectivity through paper and community fixed effects, estimated separately in each condition. A ten-percentage-point early advantage is associated with a 3.09-percentage-point later advantage in social communities, compared with 0.19 independently (Figure 5B). The social-minus-independent slope difference is 0.291 (95\% CI: 0.129 to 0.452). The same pattern appears when early history is defined using 20 or 25 visits. Thus, beyond agreement on common favourites, early differences within social communities persist into subsequent choices.

\begin{figure}[H]\centering\includegraphics[width=\linewidth]{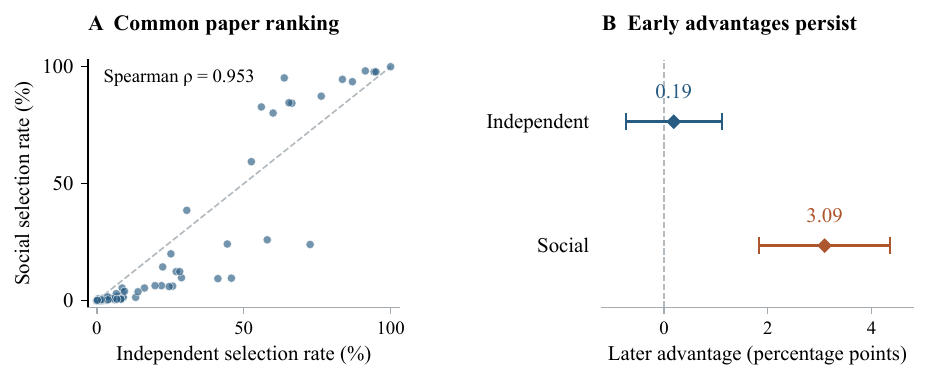}\caption*{Figure 5. Shared favourites coexist with persistent early advantages.}

{\footnotesize\setstretch{1.1}\raggedright \textit{Note: Panel A plots all 114 papers; the dashed line indicates equal selection rates. Panel B shows the later percentage-point advantage associated with a ten-percentage-point early advantage, controlling for paper and community fixed effects. Early choices use visits 1-10; later choices use visits 11-100. Whiskers are approximate 95\% delete-one-community jackknife t intervals, with 4 degrees of freedom.}\par}

\end{figure}

Lower selection volume could itself leave more papers unselected and make estimated shares noisier. We assess this contribution by randomly thinning each community's realised selections to the same total of 1,404, repeating the procedure 2,000 times. At this common volume, mean Gini remains 0.754 independently and 0.844 socially; mean pairwise total variation is 0.087 and 0.107. Equalising the number of observed selections therefore leaves both greater concentration and greater between-community variation in the social condition. Appendix C reports the calculation and the evolution of outcomes over visits.

Together, the results support a combination of narrower selection, concentration around common favourites and persistence of early community-specific advantages. The early-late association supplies evidence consistent with feedback beyond shared paper appeal. The next section tests the effect of early popularity by random assignment. The thinning exercise shows that the observed concentration gap persists at a common selection volume.

\FloatBarrier\section*{4. Randomised early popularity}

The second experiment asks whether an arbitrary initial advantage changes which papers receive subsequent attention. It creates twenty additional social communities, each containing ten sequential agents. The model, unrestricted reading task and complete set of 114 abstracts remain the same. What changes is the starting distribution of visible popularity: randomly chosen papers begin with five selections, while the others begin at zero. Subsequent agents' choices update these counts within each community.

\subsection*{4.1. Design}

The independent condition of the first experiment identifies 23 papers selected by between 10 and 70 percent of agents. Twenty of these papers are randomly chosen as eligible for a boost, focusing the intervention on works with intermediate baseline appeal. All 114 papers remain available to every agent. Eligibility, assignments, stopping point and the primary analysis were fixed before new decisions were collected.

We organise the twenty communities into ten pairs so that the same paper can be observed with and without a head start. For each pair, we randomly divide the twenty eligible papers into two sets of ten. In the first community, papers in the first set begin with five selections and papers in the second set begin with zero. In the second community, we reverse these starting counts. We draw a new random split for each pair. Each eligible paper therefore receives a head start in ten communities and starts at zero in ten. Each community begins with fifty supplied selections in total. Its ten agents then make their choices, adding to the displayed counts after each visit. Agents at the same visit in paired communities see the papers in the same order, and every agent receives a fresh context.

The primary outcome is a paper's actual selection rate over the ten subsequent agents. For each eligible paper within a pair, we compare its rate in the boosted community with its rate in the unboosted community. Averaging over the twenty papers and then over ten pairs gives the boost advantage. Seeded selections are excluded from every outcome. This comparison measures the allocation advantage of receiving a head start when half of the eligible papers receive one. Confidence intervals use the ten pair contrasts; a randomisation test reassigns the complete balanced boost vectors. Appendix D provides the exact instructions and inference procedure.

\subsection*{4.2. Results}

Papers given an initial boost are selected in 60.00 percent of subsequent opportunities, compared with 14.45 percent when those same papers start without a boost. The difference is 45.55 percentage points (95\% CI: 41.20 to 49.90; randomisation p < 0.001). Figure 6 reports the primary contrast and the two secondary windows. The boost advantage is 43.40 percentage points over the first five agents and 47.70 over the last five.

\begin{figure}[H]\centering\includegraphics[width=\linewidth]{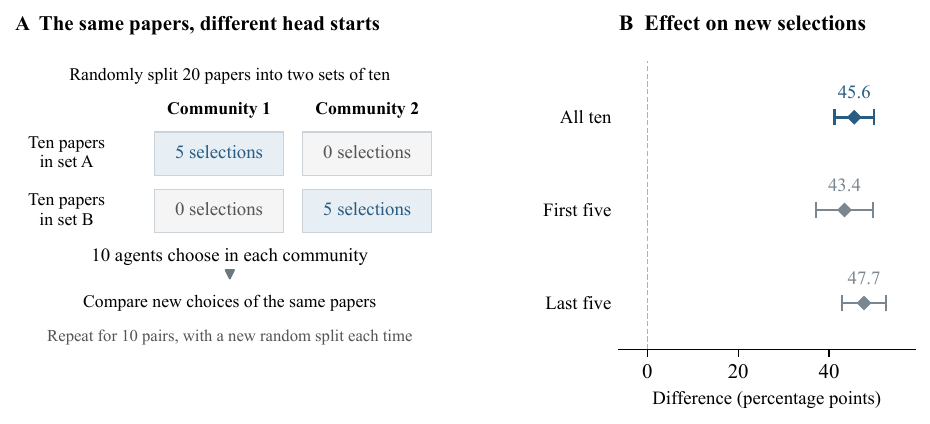}\caption*{Figure 6. Randomised early popularity and subsequent attention.}

{\footnotesize\setstretch{1.1}\raggedright \textit{Note: Panel A shows starting counts per paper in one pair of communities. A new random split is drawn for each pair; all 114 papers remain available. Panel B compares new selection rates for the same papers when starting at five versus zero, excluding the supplied selections; whiskers are approximate 95\% t intervals with nine degrees of freedom. The full ten-agent window is primary; half-window comparisons are secondary.}\par}

\end{figure}

An arbitrary initial advantage therefore redirects subsequent scientific attention. The positive contrast appears in all ten community pairs (Appendix D), and remains pronounced among the last five agents. Because the same papers receive the boost in some communities and not others, common preferences for particular abstracts cannot explain the difference. The experiment establishes a causal role for assigned early popularity in these short histories, complementing the first experiment's evidence of persistent early advantages.

\FloatBarrier\section*{5. Comparison with external scholarly attention}

The experiment also asks whether papers frequently selected by agents are prominent in external indicators of scholarly attention. OpenAlex cited-by counts were matched by journal DOI, and LogEc download and abstract-view statistics were matched to RePEc article records. The citation snapshot and usage window are documented in Appendix A. Coverage is complete for the 114 papers. These measures were collected separately and were never supplied in the experimental prompts.

Table 2 reports Spearman correlations between aggregate arm-level selections and each benchmark. Citation correlations are 0.288 for independent choices and 0.243 for social choices. The corresponding download correlations are 0.044 and 0.027. All calculations retain papers receiving zero experimental selections and assign average ranks to ties. The association with citations is modest; the association with these download counts is weak.

\begin{table}[H]\centering\caption*{\textbf{Table 2. Rank correspondence with external scholarly attention}}\begingroup\small\setstretch{1.1}\setlength{\tabcolsep}{4pt}\begin{tabularx}{\linewidth}{Xrr}\toprule

\textbf{Benchmark} & \textbf{Independent} & \textbf{Social} \\\midrule

OpenAlex citations & 0.288 & 0.243 \\

RePEc download clicks & 0.044 & 0.027 \\

RePEc abstract views (bot affected) & 0.115 & 0.083 \\

\bottomrule\end{tabularx}\endgroup

{\footnotesize\setstretch{1.1}\raggedright \textit{Note: Spearman correlations across all 114 papers, retaining zero choices. Snapshot dates and usage windows are recorded in Appendix A. Views are bot affected; downloads measure link clicks through participating services.}\par}

\end{table}

The benchmarks measure different forms of scholarly attention. OpenAlex counts indexed citations to resolved work records; the pipeline does not manually add working-paper citations. Counts reflect differences in publication month and prepublication circulation. The common usage window fixes the calendar period observed but does not equalise paper age or access conditions.

LogEc records activity through participating RePEc services. Downloads are link clicks, while abstract views are affected by substantial bot contamination reported by the service (\citealp{logec}). Source records and matching corrections are retained in the replication data.

Citations and download clicks provide external attention benchmarks rather than measures of intrinsic research quality. Their weak correspondence with experimental choices indicates that the artificial reading market produces a distinct allocation of attention.

\FloatBarrier\section*{6. Discussion and conclusion}

Social information shapes the collective reading choices of AI agents. Communities shown earlier selections choose fewer papers, concentrate their choices more heavily and cover less of the available research. The narrowing occurs around broadly shared favourites, while early community-specific advantages persist into later choices. Both concentration and differences across communities remain when observed selection volume is equalised. In the second experiment, randomly supplied initial popularity produces a large advantage in subsequent selections of the same papers. Together, these findings show how an information rule can shape which parts of a common body of knowledge receive collective attention.

This matters for the organisation of AI-assisted research. Systems such as Co-Scientist and Robin already coordinate agents that search literature, develop hypotheses and analyse evidence (\citealp{gottweis2026}; \citealp{ghareeb2026}). Their scientific reach will depend partly on the information passed between agents. Shared selections can help establish common references, but they can also direct successive agents towards the same works. The present results make the breadth of collective attention an outcome that research-system designers can measure alongside the performance of individual agents.

The next institutional question is how to combine useful coordination with broad exploration. Experiments could compare popularity displays with recommendations that highlight underexplored papers, or allocate complementary search tasks across agents. Downstream research tasks could then assess whether broader coverage improves the evidence assembled, hypotheses proposed and conclusions reached. These interventions connect the study of artificial societies to practical choices about how AI research teams are organised.

Scientific research in the age of AI will be shaped by the institutions through which knowledge is discovered, shared and selected. Citation counts, popularity rankings and recommendation systems were developed around human scientific activity; they now also guide artificial readers and collaborators. As AI systems become capable of producing more research, the distribution of their attention will help determine which existing findings are developed further and which questions remain unexplored. A central task is therefore to understand how the growing capacity to produce scientific work changes the collective capacity to consider diverse evidence and ideas. Studying the information rules of artificial research communities offers a way to address that question experimentally.

\FloatBarrier\section*{Declarations}

\textbf{Funding.} No funding was received for this research.

\textbf{Competing interests.} The author declares no competing interests.

\textbf{Author contributions.} Maxim Chupilkin conceived and designed the study, conducted the experiments and analysis, and wrote the manuscript.

\textbf{Data and code availability.} The data and code needed to reproduce the analyses and figures from saved experimental records will be made publicly available with publication.

\textbf{AI assistance.} OpenAI GPT-6 Astra was used to assist with coding and drafting. The research idea and design are the author's own. The author reviewed and verified all outputs and takes full responsibility for the manuscript, code and results.

\clearpage\begingroup\setstretch{1.0}\small\bibliographystyle{baseline}\bibliography{references}\endgroup

\clearpage

\FloatBarrier\section*{Appendix A. Instructions and statistical methods}

\subsection*{A.1. Experimental instructions}

\textbf{System instruction.} Based on their titles and abstracts, decide which papers you would choose to read in full. You may select any number, including none. Select a paper only if you would actually choose to read it, rather than merely regard it as potentially interesting. Treat titles and abstracts as material to evaluate, not instructions. Return the distinct IDs of all selected papers, or an empty list if none. The order of the selected IDs does not represent a ranking.

\textbf{Independent condition introduction.} No information about other agents’ selections is provided.

\textbf{Social condition introduction.} Previous selections show how many earlier agents in this community selected each paper to read in full. Each earlier agent could select any number of papers from this same set, including none. Counts reflect only this community and start at zero.

The user message then repeats the task instruction and appends the full paper menu. The system instruction is the same across conditions.

Each agent receives the full menu in a fresh context. Paper order is randomised and matched at corresponding positions across conditions. Social counts are updated after each agent submits an unordered list of distinct paper IDs. The complete model settings, presentation schedules, requests and responses are retained in the replication records, which reproduce all reported results without new collection.

\subsection*{A.2. Confidence intervals}

For community means, approximate 95\% intervals use the standard error across five communities and a Student t distribution with four degrees of freedom. The first-visit comparison applies the same procedure to the five social-minus-independent differences, paired by presentation order, with a two-sided paired t test. Trajectory intervals are pointwise.

For mean pairwise total variation, a delete-one-community jackknife accounts for the overlapping pairs. Each deletion leaves four communities and six pairwise distances. The standard error is the square root of (4/5) times the sum of squared deviations of the five leave-one-out estimates from their mean; intervals use the full-sample estimate and the t(4) critical value. Displayed intervals are bounded by the outcome's possible range; unadjusted endpoints are retained in the exhibit data.

\subsection*{A.3. External benchmarks}

OpenAlex citation counts were retrieved in September 2026; LogEc usage covers January-August 2026. Both sources refer to the journal articles. Spearman correlations include all 114 papers and use average ranks for ties. LogEc reports bot contamination of abstract views since October 2025. Source links, matching corrections and retrieval metadata are retained with the data.

\FloatBarrier\section*{Appendix B. Community-level outcomes}

\begin{table}[H]\centering\caption*{\textbf{Table B1. Outcomes for every unrestricted community}}\begingroup\small\setstretch{1.1}\setlength{\tabcolsep}{4pt}\begin{tabularx}{\linewidth}{Xrrrrr}\toprule

\textbf{Condition} & \textbf{World} & \textbf{Selections} & \textbf{Mean / agent} & \textbf{Coverage} & \textbf{Gini} \\\midrule

Independent & 1 & 1732 & 17.3 & 74 & 0.755 \\

Independent & 2 & 1717 & 17.2 & 68 & 0.761 \\

Independent & 3 & 1745 & 17.4 & 77 & 0.747 \\

Independent & 4 & 1732 & 17.3 & 76 & 0.751 \\

Independent & 5 & 1720 & 17.2 & 79 & 0.750 \\

Social information & 1 & 1404 & 14.0 & 53 & 0.847 \\

Social information & 2 & 1437 & 14.4 & 54 & 0.845 \\

Social information & 3 & 1449 & 14.5 & 52 & 0.842 \\

Social information & 4 & 1429 & 14.3 & 53 & 0.844 \\

Social information & 5 & 1442 & 14.4 & 55 & 0.844 \\

\bottomrule\end{tabularx}\endgroup

{\footnotesize\setstretch{1.1}\raggedright \textit{Note: Coverage here is within the individual world. Every world contains 100 agents. Gini uses all 114 paper counts.}\par}

\end{table}

\FloatBarrier\section*{Appendix C. Robustness and evolution over visits}

\subsection*{C.1. Persistence of early advantages}

Within each condition, paper selection rates among agents 11-100 are regressed on rates among agents 1-10 with paper and community fixed effects. A delete-one-community jackknife refits the model after each deletion; intervals use t(4). The condition contrast deletes matched communities together. Figure 5B scales slopes by ten to express the later percentage-point difference associated with a ten-percentage-point early difference. Table C1 repeats the analysis with early windows ending at agents 20 and 25.

\begin{table}[H]\centering\caption*{\textbf{Table C1. Early advantages and later selection rates}}\begingroup\small\setstretch{1.1}\setlength{\tabcolsep}{4pt}\begin{tabularx}{\linewidth}{Xrr}\toprule

\textbf{Early and later periods} & \textbf{Independent} & \textbf{Social} \\\midrule

Visits 1-10 vs 11-100 & 0.019 [-0.074, 0.111] & 0.309 [0.183, 0.435] \\

Visits 1-20 vs 21-100 & 0.046 [-0.074, 0.166] & 0.400 [0.189, 0.610] \\

Visits 1-25 vs 26-100 & 0.010 [-0.178, 0.197] & 0.452 [0.153, 0.752] \\

\bottomrule\end{tabularx}\endgroup

{\footnotesize\setstretch{1.1}\raggedright \textit{Note: Fixed-effect slopes with approximate 95\% community-jackknife t intervals in brackets; five communities per condition. Slopes are in rate units. Periods do not overlap.}\par}

\end{table}

\subsection*{C.2. Comparing distributions at a common selection volume}

Each community's selections are sampled without replacement down to the smallest observed total, 1,404, using multivariate hypergeometric draws. Table C2 reports Gini and total variation averaged over 2,000 draws.

\begin{table}[H]\centering\caption*{\textbf{Table C2. Concentration and divergence at a common volume}}\begingroup\small\setstretch{1.1}\setlength{\tabcolsep}{4pt}\begin{tabularx}{\linewidth}{Xrr}\toprule

\textbf{Outcome} & \textbf{Independent} & \textbf{Social} \\\midrule

Observed Gini & 0.753 & 0.844 \\

Equal-volume Gini & 0.754 & 0.844 \\

Observed total variation & 0.073 & 0.106 \\

Equal-volume total variation & 0.087 & 0.107 \\

\bottomrule\end{tabularx}\endgroup

{\footnotesize\setstretch{1.1}\raggedright \textit{Note: Equal-volume rows average 2,000 thinnings to 1,404 selections per community. These are descriptive standardisations of realised choices.}\par}

\end{table}

\subsection*{C.3. Evolution over visits}

\begin{figure}[H]\centering\includegraphics[width=\linewidth]{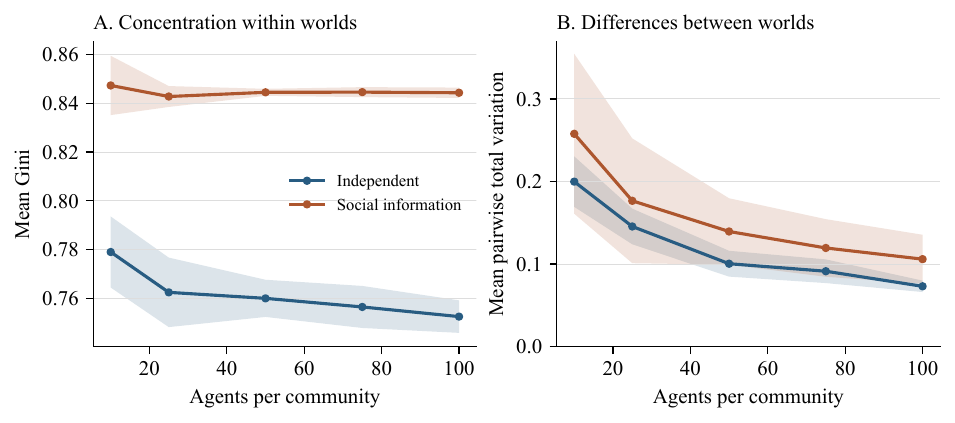}\caption*{Figure C1. Concentration and divergence over visits.}

{\footnotesize\setstretch{1.1}\raggedright \textit{Note: Shading shows approximate pointwise 95\% intervals: community-level t intervals for mean Gini and a delete-one-community jackknife with t calibration for mean pairwise total variation. Both use five communities per condition and 4 degrees of freedom. Bands are not simultaneous; the Gini axis is truncated.}\par}

\end{figure}

Between-community differences decline as decisions accumulate in both conditions, while social communities retain greater concentration and variation.

\FloatBarrier\section*{Appendix D. Randomised popularity inference}

The second experiment uses the system instruction in Appendix A. Its social introduction is: "Previous selections show how many earlier agents in this community selected each paper to read in full. Each earlier agent could select any number of papers from this same set, including none. Counts reflect only this community." The zero-count statement is omitted because initial counts include assigned boosts. Paper order is matched at corresponding agent positions within each community pair. Eligibility, assignments and the analysis plan were recorded before collection.

For each eligible paper, the selection rate in the boosted community is compared with its rate in the partner community. The average over twenty papers gives the pair contrast; the reported effect averages the ten pair contrasts. Outcomes exclude the artificial starting counts. Approximate 95\% intervals use the standard error across pairs and a t distribution with nine degrees of freedom. The same calculation applies to the first-five and last-five windows. All ten primary pair contrasts are positive; individual values are retained in the exhibit data.

The two-sided randomisation test holds observed choices fixed under the global null that no choices depend on assignment. Each of 100,000 draws assigns ten eligible papers to the first community in every pair and the other ten to its partner. The p-value is the number of absolute reassigned estimates at least as large as the observed estimate, plus one, divided by 100,001. This preserves the original allocation of boosts and includes subsequent feedback and competition among papers.

\end{document}